\documentclass[letterpaper, 10 pt, conference]{ieeeconf}  

\IEEEoverridecommandlockouts                              

\usepackage{times} 
\usepackage[pdftex]{graphicx}
\usepackage[usenames,dvips,pdftex]{color}
\usepackage{wrapfig}
\usepackage{amsmath, amsfonts}
\usepackage{amssymb}
\usepackage{longtable}
\usepackage{wrapfig}
\usepackage{cite}
\usepackage{units}
\usepackage{url}
\usepackage{algorithm}
\usepackage[caption=true]{subfig}
\usepackage{pdfpages}
\usepackage{epstopdf}
\usepackage{caption}
\usepackage[noend]{algpseudocode}
\usepackage{array,multirow}
\usepackage{hhline}
\usepackage{booktabs}
\usepackage{stackengine}
\usepackage[percent]{overpic}
\usepackage{relsize}

\newcommand{\matr}[1]{\boldsymbol{#1}}	
\newcommand{\set}[1]{\boldsymbol{#1}}	

\newcommand{\domain}[1]{\mathbb{#1}}
\newcommand{\Epsilon}{\mathcal{E}}

\newcommand{\cnn}[1]{\emph{#1}}	

\makeatletter
\ifdefined \r@draft \newcommand{\cblue}[1]{\textcolor{blue}{#1}} 
\else \newcommand{\cblue}[1]{\iffalse{#1}\fi}
\fi
\makeatother

\newcommand{\Figure}[1]{Fig.~\ref{#1}}

\newcommand{\Table}[1]{Table~\ref{#1}}
\newcommand{\Section}[1]{Sec.~\ref{#1}}

\usepackage{xcolor}
\usepackage{cancel}

\newcommand{\approve}[1]{#1}

\newcommand{\specialcell}[2][c]{%
  \begin{tabular}[#1]{@{}c@{}}#2\end{tabular}}

\title{\LARGE \bf
CNN for Very Fast Ground Segmentation in Velodyne LiDAR Data
}

\author{Martin Velas, Michal Spanel, Michal Hradis and Adam Herout
\thanks{All authors are with with the Faculty of Information Technology, Brno University of Technology, Czech Republic {\tt\small \{ivelas$\vert$spanel$\vert$ihradis$\vert$herout\} at fit.vutbr.cz}}
\thanks{*This work has been supported by the Artemis JU grant agreement ALMARVI (no. 621439), the TACR Competence Centres project V3C (no. TE01020415), and the IT4IXS – IT4Innovations Excellence project (LQ1602).}
\thanks{**We would like to thank Mingfang Zhang for sharing her source code \cite{Zhang15}, making evaluation of both approaches feasible.}
}

\begin{document}

\maketitle
\thispagestyle{empty}
\pagestyle{empty}

\begin{abstract}
This paper presents a novel method for \emph{ground segmentation in Velodyne} point clouds. We propose an encoding of sparse 3D data from the Velodyne sensor suitable for training a \emph{convolutional neural network (CNN)}. This general purpose \approve{approach} is used for segmentation of the sparse point cloud into ground and non-ground points. \approve{The LiDAR data are represented as a multi-channel 2D signal where the horizontal axis corresponds to the rotation angle and the vertical axis the indexes channels (i.e. laser beams). Multiple topologies of relatively shallow CNNs (i.e. 3-5 convolutional layers) are trained and evaluated using a manually annotated dataset we prepared.} The results show significant improvement of performance over the state-of-the-art method by Zhang et al. in terms of \emph{speed} and also minor improvements in terms of accuracy.
\end{abstract}

\section{Introduction}
\begin{figure}[t]
	\includegraphics[width=\linewidth]{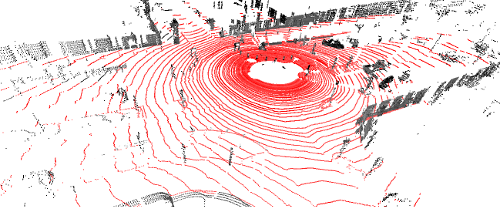}
	\caption{Expected segmentation of Velodyne LiDAR point cloud into the sets of ground (red) and not-ground (grey) points.\label{fig:ground-det-example}}
    \vspace{-1em}
\end{figure}

Recent development in exploration and $3$D mapping of the environment surrounding a mobile robot aims at techniques which capture semantic information besides the simple geometrical properties. The analysis of scene dynamics was successfully used in the task of object detection (pedestrians, cars, bicycles, ...) \cite{Wang12} and by filtering out moving objects; the $3$D maps capturing only static parts of the scene can be built \cite{Jiang16}. Such maps are useful for the localization or motion planning, where the records of objects moving in the past are undesirable.  
Successful methods for the \emph{detection and tracking of moving objects (DATMO)} assume that sensors are used in the way that only the objects (static or dynamic) are captured~\cite{Tanzmeister14}, or that the ground can be detected (see \Figure{fig:ground-det-example}) and filtered out in the preprocessing stage \cite{Li14, Choi13, Wojke12, Asvadi15, Mertz13}. 
For these purposes, we intend to reliably and efficiently \emph{segment the data to ground/non-ground} parts. \approve{To be more specific, we consider the ground to be every surface traversable by common moving objects (pedestrians, cars, bikes, etc.).}

In these DATMO systems, the ground detection is ty\-pi\-cal\-ly based on primitive features with low discriminative capabilities. A state of the art technique for robust ground segmentation by Zhang et al. \cite{Zhang15} achieves good results in terms of accuracy by building a Markov Random Field (MRF) and inference using the Loopy belief propagation. Unfortunately, the robustness of this method is achieved by compromising its time efficiency (over $2\,\mathrm{min}$ per frame).

\approve{One of the common source of \emph{LiDAR} (Light Detection And Ranging) data -- the \emph{Velodyne} sensor -- captures the full $3$D information of the environment} comparing to the simple range finders providing only information about occupancy in a certain height around the robotic platform. Currently the most powerful model HDL-$64$E covers full $360^{\circ}$ horizontal field and $26.8^{\circ}$ vertical field of view and with up to $15$\,Hz frame rate captures over $1.3$\,M of points per second.
This sensor scans the surrounding area by $64$ rotating laser beams where each beam produces one \emph{ring} of $3$D points (red circles in \Figure{fig:ground-det-example}).

Since the breakthrough done by AlexNet \cite{AlexNet}, the attractiveness of \emph{Convolutional Neural Networks (CNNs)} has rapidly grown and this model was successfully used for many computer vision tasks including image classification, object detection, face recognition, semantic segmentation \cite{Long15}, etc. In this work, we deployed convolutional neural networks for the task of \emph{ground segmentation} in sparse Velodyne point cloud data. We designed multiple networks with shallow topologies ($3$-$5$ convolutional layers) fulfilling the requirements for robustness and accuracy, we trained and evaluated them by using the prepared and hand-annotated dataset.

The main contributions of this work are the following:
\begin{itemize}
	\item we show that the sparse 3D \emph{LiDAR data can be encoded} into a multi-channel 2D signal (analogous to HHA encoded range images \cite{Gupta14} or \approve{LiDAR data encoding in vehicle detection task \cite{BoLi16}}) and processed by convolution neural network,
	\item an approach to \emph{CNN for ground segmentation} in Velodyne point clouds which outperforms current state of the art in accuracy and time performance.
\end{itemize}

Besides this, we developed a \emph{semi-automatic ground annotation} tool and we annotated a part of the KITTI tracking dataset. All the source code of this annotation tool, preprocessing of LiDAR point clouds, design and configuration of trained convolutional networks, as well as annotated ground truth data are publicly available\footnote{https://www.github.com/to-be-added-after-acceptance}.

\section{Related work}

\approve{As mentioned above, we define the ground as a surface traversable by common moving objects. A similar traversability estimation for an outdoor robot was performed using geometric features (extracted from stereo-vision) and texture features (from RGB images) \cite{Dongshin06}. By clustering, the labels are assigned to the parts of surrounding environment. Compared to our approach, this method requires explicit feature specification and different type of input data -- stereo RGB vision, IMU and motor current sensor.}

\approve{Convolutional networks were deployed for learning rich descriptors of RGB-D data \cite{Gupta14} useful for the per-pixel object detection. The input of network encodes horizontal disparity (equivalent to the range), height and normals angle. Our work proposes a similar type of encoding suitable for processing the sparse LiDAR data. Since the normals can not be robustly estimated in these data rather the angles are not used.}

\approve{Simultaneously with our work, Baidu research team \cite{BoLi16} proposed similar encoding of sparse LiDAR data into $2$D matrices for the vehicles detection by convolutional neural networks. The difference of proposed encoding in out work is polar bin aggregation of LiDAR points (described in \Section{sec:sparse-to-mat}) to improve the stability of prediction.}

Many DATMO (detection and tracking of moving objects) methods segment and filter out the ground measurements from LiDAR data in a preprocessing stage \cite{Li14, Choi13, Asvadi15, Mertz13, Petrovskaya08RSS, Wojke12, Petrovskaya08ISER}. \approve{Primitive features with low discriminative capabilities are used: mean or variance of measured height in a certain small area, or changes in the elevation between the rings in Velodyne data.}

More traditional DATMO methods operate over data from simple laser rangefinders~\cite{Tanzmeister14}, assuming the measurements provided by LiDAR positioned approximately parallel to the ground surface capturing only the upright (moving and/or static) objects and not the ground. Over such data, the occupancy grid can be built and detection of movement is performed by particle filtering.

\begin{figure}
	\includegraphics[width=\linewidth]{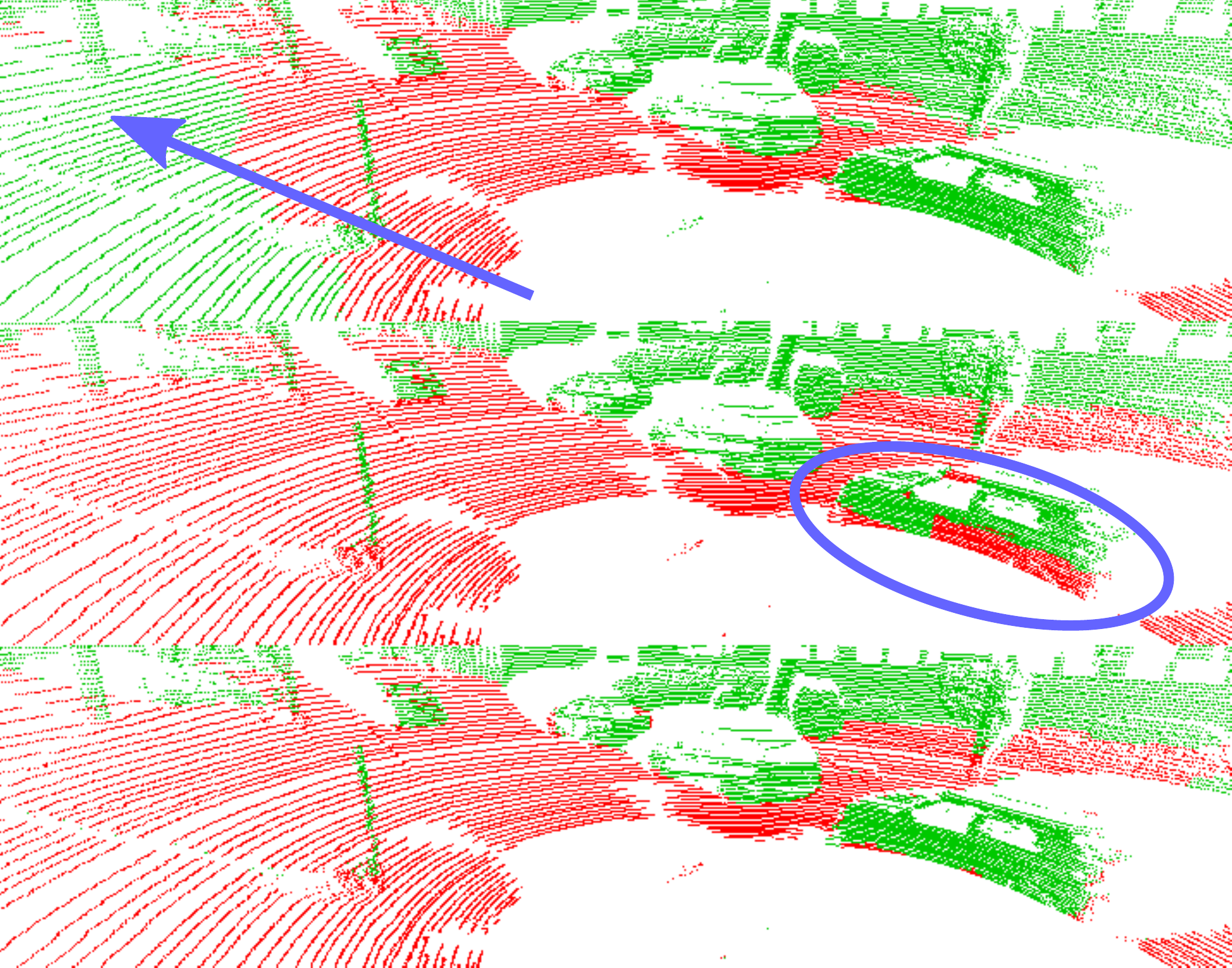}
    \caption{Different results of ground segmentation methods.
    \textbf{top:} Simple height thresholding can not deal with terrain elevation;
    \textbf{middle:} Loopy belief propagation \cite{Zhang15} produces incorrect results when objects are close to the sensor; 
    \textbf{bottom:} our method.}
    \label{fig:ground-labeling}
    \vspace{-1em}
\end{figure}

When data from multiple laser sensors including Velodyne $3$D LiDAR are fused \cite{Mertz13}, building the occupancy grid starts to be an issue, since the sensors cover a significantly larger area including the ground. The ground measurements must be recognized and filtered out in order to build a valid occupancy grid representing free space, the space occupied by obstacles and currently unobserved areas. For sake of effectivity, the authors \cite{Mertz13} selected a computationally inexpensive approach, when all measurements within a certain height range are considered to be ground. Besides the sensitivity to selection of optimal thresholds, the robustness/repeatability of such approach is far from the optimal (see \Figure{fig:ground-labeling}).

The motion detection generalized to motion field estimation \cite{Li14} in a polar grid benefits from the large area covered by the Velodyne LiDAR scanner. The same preprocessing step as in the previously mentioned work -- i.e. the ground detection and filtering -- is performed as well. Using the simple thresholding, this method shares the same disadvantages. The areas (polar grid cells) fulfilling at least one of the following conditions are considered to be  ground: the average height fits an exactly defined range, the standard deviation of the height is below a certain threshold, or the difference between the minimal and the maximal height inside the cell is below another threshold.
A very similar approach with just small modifications was used by Asvadi et al. \cite{Asvadi15} in a DATMO system operating over a regular orthogonal grid. The area within one grid cell is considered to be ground if both the mean height and the standard deviation of the heights fit below a predefined threshold.

Other approaches analyse changes in the elevation in order to segment ground in Velodyne LiDAR scans \cite{Petrovskaya08RSS, Wojke12, Petrovskaya08ISER}: each vertical slice, consisting of all the points captured at exactly the same moment by all laser rays, is analysed separately. Three points $A$, $B$, $C$ from adjacent rings form two vectors $\overrightarrow{AB}$ and $\overrightarrow{BC}$. If the dot product of these (normalized) vectors is above a certain threshold, the significant change of elevation -- the breakpoint -- is found. The breakpoint forms the border between the ground part (points between the sensor and the breakpoint) and an obstacle (points behind the breakpoint). Besides the lack of robustness, this approach does not allow to reason about the space behind the first obstacle where the ground can be observed again.

\begin{figure}
	\includegraphics[width=\linewidth]{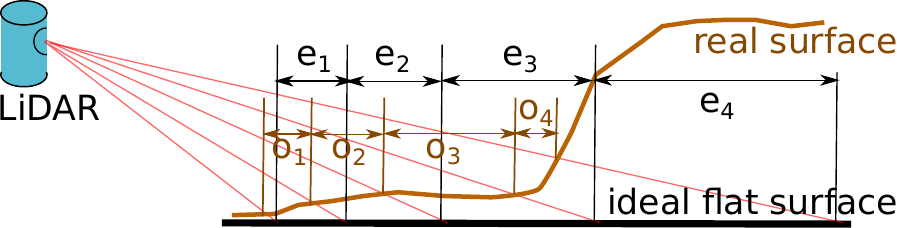}
    \caption{\label{fig:range-diffs}Ground detection by comparison of expected range difference $e_i$ with observed difference $o_i$. Since $e_4 - o_4 > th$, the border between the obstacle and the ground is found \cite{Choi13}.}
    \vspace{-1em}
\end{figure}

Analysis of the range difference between two adjacent Velodyne rings (\Figure{fig:range-diffs}) was also used for the ground segmentation \cite{Choi13} in LiDAR data. On an ideal flat horizontal surface, the expected range difference $e_i$ between two adjacent rings can be computed, assuming the height and the vertical angle of each laser beam is known. This range difference decreases with increasing elevation of the surface. At the ideal vertical obstacle, this difference becomes zero.

\approve{Besides the previously mentioned DATMO methods, the ground detection and filtering plays important role in the point cloud registration by scan segments matching \cite{Douillard12}. As a preprocessing step, the ground points are also detected by thresholding the mean and the variance of vertical height withing the cells of voxel grid \cite{Douillard10}.}

The lack of accuracy and robustness of previously mentioned methods, mostly caused by the fixed thresholding of simple features with low discriminative power, was overcome by the inference in \emph{Markov Random Field (MRF)} \cite{Zhang15}. Although the introduced $3$D volumetric grid is built in a similar way as the $2$D occupancy grids are, by estimation of the slope in each vertical slice, the final segmentation to ground/obstacle is not made directly. First, based on the slope detected, the points are categorized as unknown, probably ground, probably obstacle, and probably obstacle borders. This categorization implies the initial cost assigned to each volumetric element of the regular $3$D polar grid. The key improvement is done by Loopy Belief Propagation inference in order to estimate ground height within a certain region. All measurements within this region with a smaller height are considered to be the ground points. The rest is classified as non-ground. Unfortunately, the robustness of this method is achieved by compromising its time efficiency. In our experiments with the original MATLAB implementation kindly provided by the authors, the processing of a single Velodyne HDL-$64$E frame takes approximately $145$ seconds.

The key improvement of our method is reduction of the time required for the ground segmentation process to the fraction of the time required by Loopy Belief Propagation \cite{Zhang15} while obtaining outperforming results in terms of accuracy as well. Processing of a single Velodyne frame by our \cnn{L05+deconv} network takes $140$\,ms on average using only CPU. By using GPU (GeForce GTX 770), the processing time is further reduced to $7$\,ms per frame.

\section{Proposed Ground Segmentation Method}
\label{sec:method}

The goal of our method is to assign a \emph{binary label ground/not-ground} \approve{\eqref{eq:point-assignment}} to each $3$D point \approve{$\matr{p}\in\set{P}$} measured by the LiDAR sensor. These point cloud's elements are represented by $3$D coordinates, originating at the LiDAR sensor position, accompanied by the laser intensity reading and the ring ID which identifies the source laser beam which was used to measure the point \approve{$\matr{p} = \left[p_x, p_y, p_z, p_i, p_r\right]$. Since we do not assign the ground label to each LiDAR point separately, we solve the assignment \eqref{eq:cloud-assignment} of binary labels to all the points jointly.}
\begin{eqnarray}
	g: \set{P} \rightarrow \{0,1\} \label{eq:point-assignment}\\
    G: \domain{P} \rightarrow \{0,1\}^{\vert \set{P} \vert}, \quad \set{P} \in \domain{P} \label{eq:cloud-assignment}
\end{eqnarray}

\subsection{Encoding Sparse $3$D Data Into a Dense $2$D Matrix} 
\label{sec:sparse-to-mat}

\begin{figure}[t]
	\includegraphics[width=\linewidth]{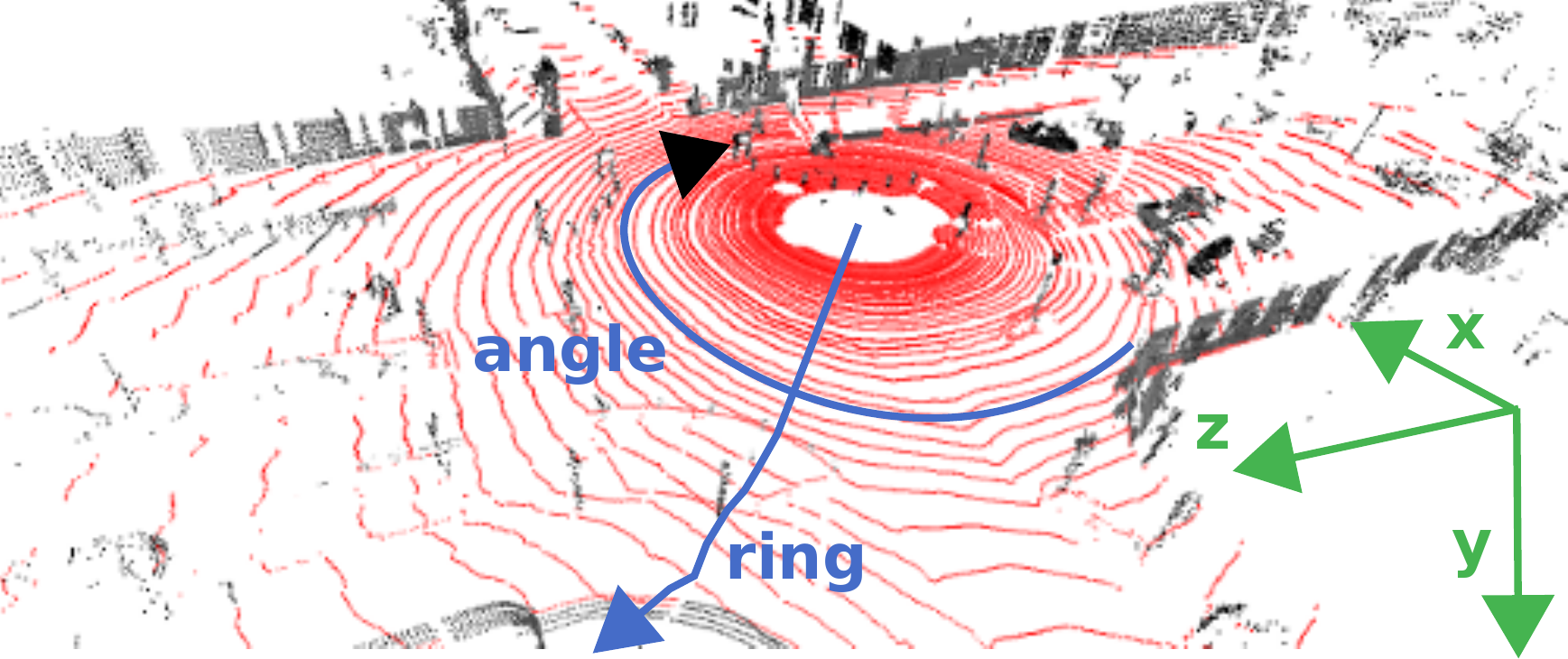}
	\vspace{-2em}

	$$\mathbb{\mathlarger{\mathlarger{\mathlarger{\Downarrow}}}}$$

	\vspace{-0.6em}
	\includegraphics[width=\linewidth]{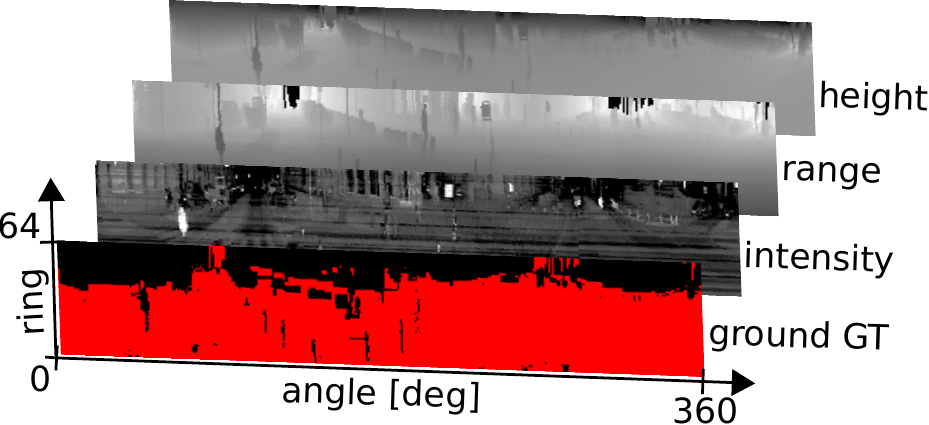}
	\caption{Transformation of the sparse Velodyne point cloud into the multi-channel dense matrix. Each row represents measurements  of a single laser beam done during one rotation of the sensor. Each column contains measurements  of all 64 laser beams captured at a specific rotational angle at the same time.\label{fig:cloud-to-matrix}}
    \vspace{-1em}
\end{figure}

In order to process the Velodyne LiDAR data by a convolutional neural network, we \emph{encode} \eqref{eq:cloud-encoding} the original \emph{sparse point cloud} $\matr{P}$ into a multi-channel \emph{dense matrix} $\matr{M}$. The original $3$D data are treated as a $2$D signal in the domain of the ring (the ID of the source laser beam) and the horizontal angle, as illustrated in \Figure{fig:cloud-to-matrix}. The size of the resulting matrix $\matr{M}$ depends on the number of rings in the LiDAR frame (i.e. number of laser beams used) and the sampling rate $R$ of the horizontal angle. In our experiments, we used Velodyne LiDAR HDL-$64$E with $64$ rays and resolution $R=1^\circ$.
\begin{equation}
	G(\set{P}) = \tilde{G}(\matr{M}), \quad \matr{M} = \Epsilon(\set{P}) \label{eq:cloud-encoding}
\end{equation}

First, the point cloud is \approve{aggregated into the polar bins $\set{b_{r,c}}$} \eqref{eq:polar-bin} analogous to our previous work \cite{Velas16}. All the points assigned to the same bin share the same ring ID $r$ (points captured by the same laser beam) and fit into the same polar cone $c = \varphi(p)$ \eqref{eq:polar-cone}, computed according to the horizontal angle of the point. \approve{Each polar bin is encoded into the element $\mathbf{m_{r,c}}$ of matrix $\matr{M}$ in the $r$-th row and $c$-th column \eqref{eq:bin-encoding-1}}. Since multiple points fall into the same bin \approve{(the horizontal representation of our encoding is coarser than original Velodyne resolution)}, a single representative is found as the average \eqref{eq:bin-encoding-2}. Moreover, since the horizontal index in the matrix $\matr{M}$ encodes the rotational angle in the $3$D horizontal $XZ$ plane, we can reduce the number of channels by replacing $XZ$ coordinates $p_x, p_z$ by \approve{depth value $d = \left\lVert p^x,p^z\right\rVert_2$ (range) without loss of any information.}
\begin{align}
	\matr{m}_{r, c} &= \varepsilon(\set{b}_{r, c}) \label{eq:bin-encoding-1}\\
	\varepsilon(\set{b_{r,c}}) &= \dfrac{\mathlarger{\sum}\limits_{\matr{p}\in \set{b_{r,c}}} \left[p^y, \left\lVert p^x,p^z\right\rVert_2, p^i\right]}{\vert \set{b_{r,c}} \vert} \label{eq:bin-encoding-2}\\
    \set{b_{r,c}} &= \{ p \in \set{P} \,\vert\, p_{r} = r \wedge \varphi(p) = c \} \label{eq:polar-bin}\\
    \varphi(p) &= \left\lfloor \dfrac{\mbox{atan}(\frac{p_z}{p_x}) + 180}{\frac{360}{R}} \right\rfloor \label{eq:polar-cone}
\end{align}

In case of empty bins (e.g. no measurement exists in this area due to the sensor limits), the value in the matrix $\matr{M}$ is linearly interpolated from the neighbourhood.

\subsection{Training Dataset}
\label{sec:data-prep-training}

The most serious issue in development of the proposed system was the lack of training data,  especially missing annotations of ground data in the Velodyne scans. The developement of KITTI Semantic Segmentation dataset\footnote{http://www.cvlibs.net/datasets/kitti/eval\_semantics.php} is still a work in progress and only small subsets are available for now. The only annotations relevant to our task were created by Richard Zhang~\cite{RZhang15} in his work on semantic segmentation of urban scenes. However, in this work, the LiDAR point clouds were used only as a supplementary data and annotations were prepared for RGB camera images in the first place. These annotations were probably back-projected into the LiDAR frames and spread across consequent frames what caused serious inaccuracies in the ground annotations and made these data unfit for our training and testing.

\begin{figure}[t]
	\centering
	\includegraphics[width=0.99\linewidth]{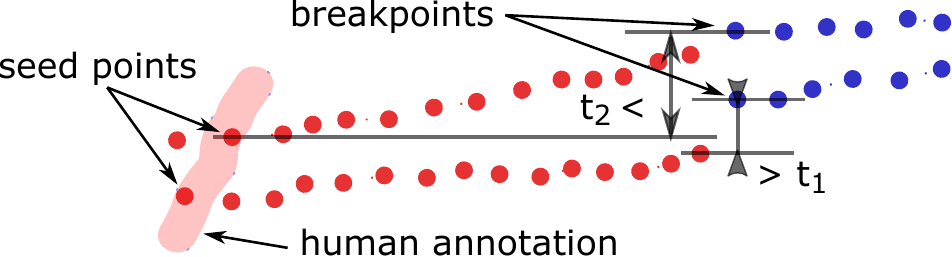}
	\caption{Flooding the human made annotation from seed points along the ring. The ground points are red. When the breakpoint is found (first of blue not-ground points), the flooding is stopped.}
    \label{fig:ann-tool}
    \vspace{-1em}
\end{figure}

\begin{figure*}[t]
	\centering
	\includegraphics[width=0.99\textwidth]{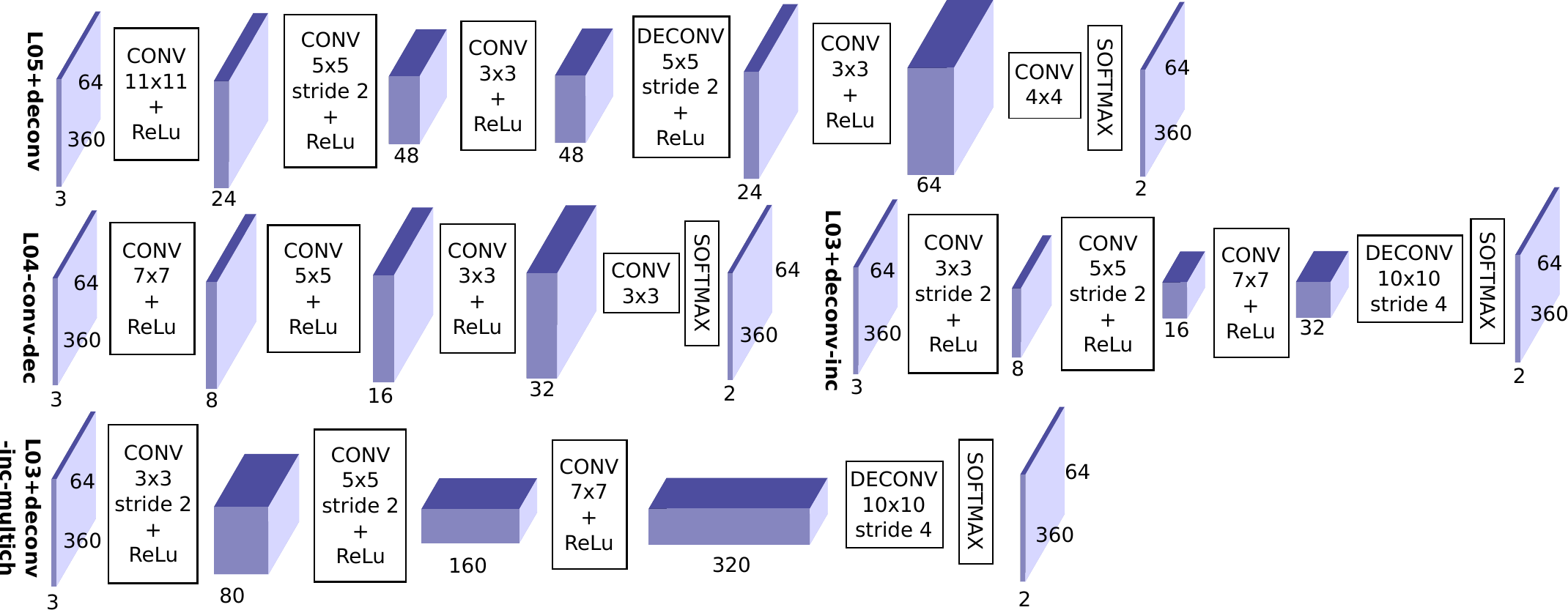}
    \caption{\label{fig:cnn-topologies} Topology of the four proposed CNNs including dimensions of intermediate data blobs (blue blocks) and the number of channels below each blob. \approve{\emph{L05+deconv} consists of $5$ convolutional layers plus single deconvolution to restore the original frame width/height. \emph{L04-conv-dec} process the input frame by $4$ convolutional layers with decreasing size ($7, 5, 3, 3$) of convolution kernel. In \emph{L03+deconv-inc}, $3$ convolutional layers with increasing kernel size are used. Deconvolution is used to restore original frame size in both this topology and in \emph{L03+deconv-inc-multich} where the number of output channels are significantly larger comparing with other networks.} Note: if the stride parameter $N$ is set in (de-)colutional layers, the width and height of the output blob is (larger or) smaller $N$-times.}
    \vspace{-0.5em}
\end{figure*}

Therefore we prepared a \emph{semiautomatic tool for ground annotation} in $3$D Velodyne data\footnote{https://www.github.com/to-be-added-after-acceptance}. Using a pen-like drawing tool, the user highlights certain ground points as ground seed points $\matr{p}^s$. From these points, the annotation \emph{automatically floods} along the ring until a breakpoint $\matr{p}^b$ is found (see \Figure{fig:ann-tool}). The breakpoint is defined as the first point where the height difference with respect to the previous point  $\vert p^b_y - p^{i-1}_y \vert > t_1$, or with respect to the seed point $\vert p^b_y - p^s_y \vert > t_2$, is above a respective threshold. When annotating the dataset, we found the values $t_1 = 3\,\mathrm{cm}$ and $t_2 = 7\,\mathrm{cm}$ work best saving annotator's time and reducing manual changes.

Using this tool, we prepared \emph{\approve{accurate} annotations of ground} in $3$D LiDAR data for a subset of KITTI Tracking Dataset -- the same data as was annotated by \cite{RZhang15} in RGB images. The subset consists of $8$ data sequences taken at different places of urban and suburban environment. In total, there are $252$ frames captured in $1\,\mathrm{s}$ interval. We randomly split these frames into training and evaluation set in $70:30$ ratio.

Since the amount of available annotated data is quite small, we prepared automatic artificial annotations of the rest of the KITTI Tracking Dataset ($19$\,k frames) by thresholding simple features like the mean and the variance of height, the distance and the elevation differences between rings as used in the previous works \cite{Li14, Choi13, Asvadi15, Mertz13, Petrovskaya08RSS, Wojke12, Petrovskaya08ISER}. These artificial annotations are used for CNNs pretraining whose resulting parameters are used as initial weights of convolutional kernels for further training on human annotated data.

\approve{We also tried to use data augmentation and generate the artificial 3D LiDAR data automatically. Unfortunately this approach proved to be infeasible, since the available 3D models are not detailed enough, lack the structure information and substitute this 3D structure (trees, bushes, curbs, etc.) by texturing the flat surfaces.}

\subsection{Topology and Training of the Proposed Networks}

Because of the small amount of annotated training data, we used \emph{shallow CNN architectures} only. The networks are fully convolutional. They consists of convolution and deconvolution layers with ReLU non-linearities. Gradient descent is used as the optimization method for the training. The most interesting and successful topologies we experimented with are presented in \Figure{fig:cnn-topologies}.

The multi-channel matrix $\matr{M}$, obtained by the encoding described in \ref{sec:sparse-to-mat}, is the input of all the proposed networks. The probability of being ground point \approve{$p_g = p(g(\matr{p})=1)$} is estimated for each pixel of this matrix. Therefore, the output of all networks has the same size as the input matrices except the number of channels. The output channels represent probabilities $p_g$ and $1-p_g$ since the softmax function is applied to the output.

The presented architectures differ in the type and number of layers used, dimension of convolutional kernels, and in the number of channels within each layer. Deconvolutional layers (previously also used in semantic segmentation \cite{Long15}) were used in $3$ of $4$ presented topologies including the best topology \cnn{L05+deconv} which performs best in our experiments. In topologies \cnn{L05+deconv} and \cnn{L04-conv-dec}, the size of the convolutional layers is decreasing when compared to the other two topologies. The effect of a significantly larger number of intermediate output channels is evaluated for topology \cnn{L03+deconv-inc-multich}. See \Figure{fig:cnn-topologies} for more details.

The input of the CNN, which is prepared as described in \Section{sec:sparse-to-mat} Eq. (\ref{eq:cloud-encoding}-\ref{eq:polar-cone}), is normalized and rescaled \eqref{eq:normalization}. This applies only to the depth $d$ and the height $p_y$ channels, since the intensity values of Velodyne sensor are already normalized to interval $(0;1)$. In our experiments, the normalization constant is set to $H=3$, since in usual scenarios, the Velodyne model HDL-$64$E captures vertical slice approximately $3\mathrm{m}$ high.
\begin{equation}
	\overline{p_y} = \frac{p_y}{H}, \qquad \overline{d} = \log{(d)} \label{eq:normalization}
\end{equation}

\approve{We applied this logarithmic rescaling for the depth channel so that range differences between consequent rings are approximately the same for flat surfaces both close and far from the sensor. The rescaling should suppress the differences between the rings due to varying distance from the sensor and highlight the differences caused by the structure of observed scene -- i.e. the obstacles (illustrated in \Figure{fig:range-diffs}). In the similar manner, the horizontal disparity was previously used as an input of a convolutional network instead of using range value directly \cite{Gupta14} what results in a normalization similar to ours.}

\section{Experiments}

The proposed convolutional networks were implemented, trained and evaluated using \emph{Caffe}\footnote{http://caffe.berkeleyvision.org/} deep learning framework. Both the human annotated dataset and the automatically annotated dataset were used for training and pre-training the proposed networks, respectively.
We compared the results of our CNNs with the results of the robust state-of-the-art method \cite{Zhang15} (using the original MATLAB implementation shared by the authors). 
It is necessary to mention one limitation of the Zhang's method. Because dimensions of the polar grid need to be set, the maximal range from the sensor is limited. In the experiments we used the $60$m limit by default (and the $30$m limit in the time performance test). In order to make fair evaluation, we computed the accuracy of our method for both the maximal range set to $60$m (same conditions as for \cite{Zhang15}) and the unlimited range (to illustrate behavior for more distant measurements). Also, since the Zhang's method has no threshold/parameter for tuning the false positives to false negatives ratio, only a single precision/recall value can be computed instead of the whole PR curve.

\begin{figure}[t]
	\centering
	\includegraphics[width=0.99\linewidth]{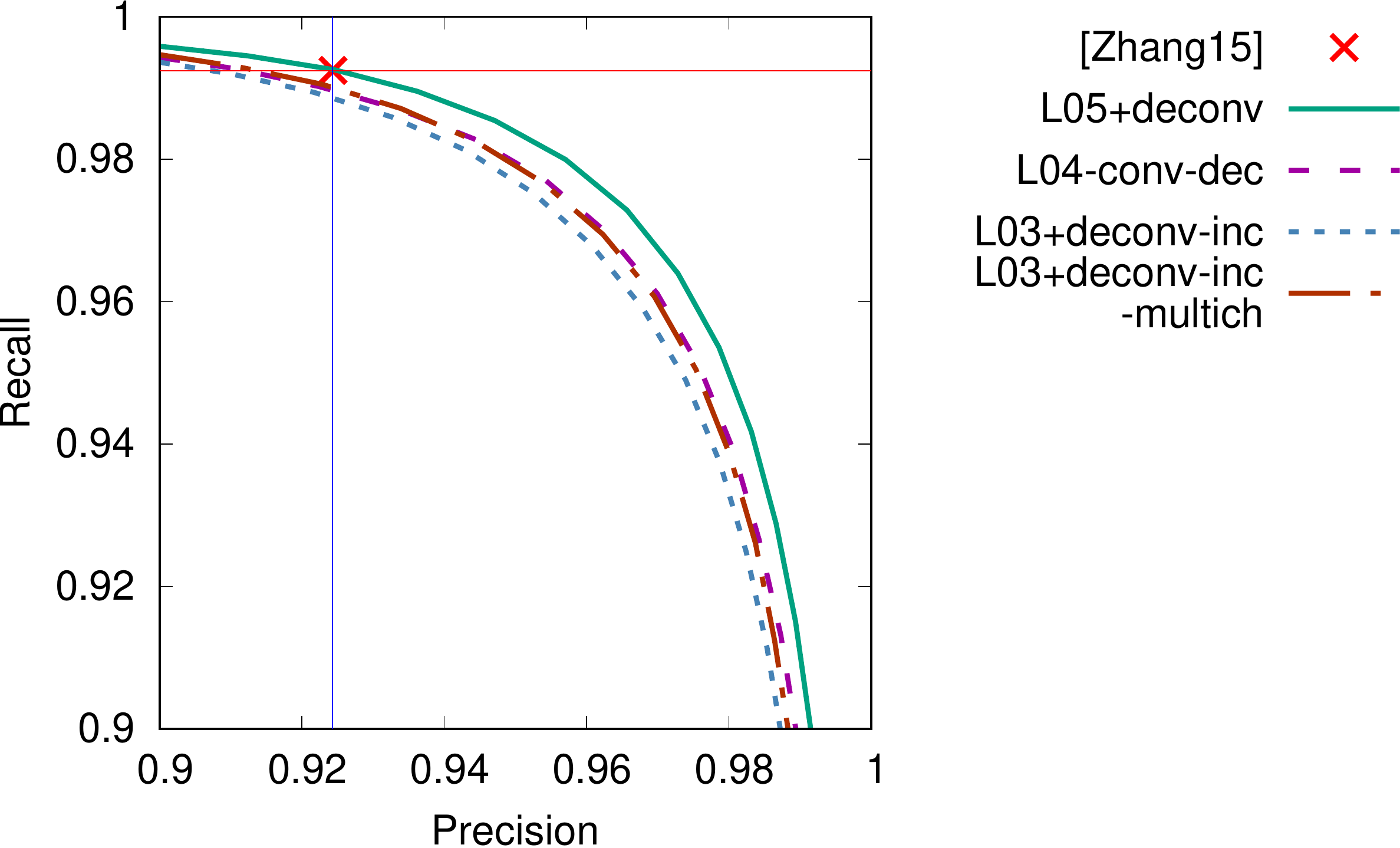}
	\caption{The accuracy of the proposed networks and the reference method \cite{Zhang15} for comparison. See \Table{tab:f-score} for numerical results.}
    \label{fig:pr-all-nets}
\end{figure}

\Figure{fig:pr-all-nets} shows the comparison of different networks with the reference method~\cite{Zhang15}. The results are also summarized in \Table{tab:f-score} by means of the average precision and F-score as the metrics of accuracy. All networks were pre-trained using the automatically annotated data, trained and evaluated using the human annotated data and only the points within the range of $60$m were taken into account.

The results (\Figure{fig:pr-all-nets} and \Table{tab:f-score}) show that the accuracy is quite similar for different network topologies. Better accuracy is achieved with the networks where the size of convolution kernels decreases (\cnn{L05+deconv} and \cnn{L04-conv-dec} CNNs) and also with larger networks. The accuracy of \cnn{L05+deconv} network is also slightly higher compared to the reference method~\cite{Zhang15}. Preserving the same recall we were able to achieve $0.5\%$ better precision and vice versa: $0.1\%$ higher recall while preserving the same precision. Also, since our method enables balancing FP:FN ratio, we were able to find an optimal operating point yielding better F-score.

\begin{table}[t]
	\centering
    \def\arraystretch{1.3}
	\begin{tabular}{@{}l|c|c|c|c@{}}
    	\toprule
		 & \textbf{AP} & \specialcell{\textbf{Precision*}\\recall=.992} & \specialcell{\textbf{Recall*}\\prec=.924} & \specialcell{\textbf{Best}\\\textbf{F-score}} \\\hline\hline
		 [Zhang15]              & -              & 0.924          & 0.992          & 0.957 \\\hline
		 \textbf{L05+deconv}    & \textbf{0.996} & \textbf{0.929} & \textbf{0.993} & \textbf{0.969} \\\hline
		 L04-conv-dec           & 0.995          & 0.914          & 0.990          & 0.966 \\\hline
		 L03+deconv-inc         & 0.994          & 0.910          & 0.989          & 0.964 \\\hline
		 L03+deconv-inc-multich & 0.995          & 0.916          & 0.990          & 0.966 \\\hline
         \bottomrule
	\end{tabular}
	\caption{Average precision (area under the PR curve), precision, recall and the best F-score of the proposed networks compared to \cite{Zhang15}. \textbf{*}The precision (and the recall) were estimated for points where recall (and precision respectively) is the same as the results of \cite{Zhang15} (also displayed in \Figure{fig:pr-all-nets} by red and blue line). The best F-score is taken as the highest value of harmonical average of precision and recall within the whole PR curve.}
    \label{tab:f-score}
    \vspace{-1em}
\end{table}

\begin{table}[t]
	\centering
    \def\arraystretch{1.3}
	\begin{tabular}{@{}l|r|r@{}}
    	\toprule
		 & \textbf{CPU only [ms]} & \textbf{with GPU [ms]} \\\hline\hline
		 \hspace{0.5em} L05+deconv     		& 139 & 7.0 \hspace{0.5em} \\\hline
		 \hspace{0.5em} L04-conv-dec   		&  90 & 3.2 \hspace{0.5em} \\\hline
		 \hspace{0.5em} L03+deconv-inc         &   8 & 1.2 \hspace{0.5em} \\\hline
		 \hspace{0.5em} L03+deconv-inc-multich & 355 & 6.9 \hspace{0.5em} \\\hline
         \bottomrule
	\end{tabular}
	\caption{\label{tab:time-performance}Performance comparison of the proposed networks in terms of speed. The average processing time per single Velodyne LiDAR HDL-$64$E frame is presented. The mini-batches of size 4 were used (i.e. 4 frames were processed in parallel).}
    \vspace{-1em}
\end{table}

\captionsetup[subfigure]{labelformat=empty}
\newcommand{\subfigLabelPadding}{2.6em}
\begin{figure*}[t]
	\subfloat[\hspace{\subfigLabelPadding}L05+deconv]{
    	\includegraphics[width=0.32\linewidth]{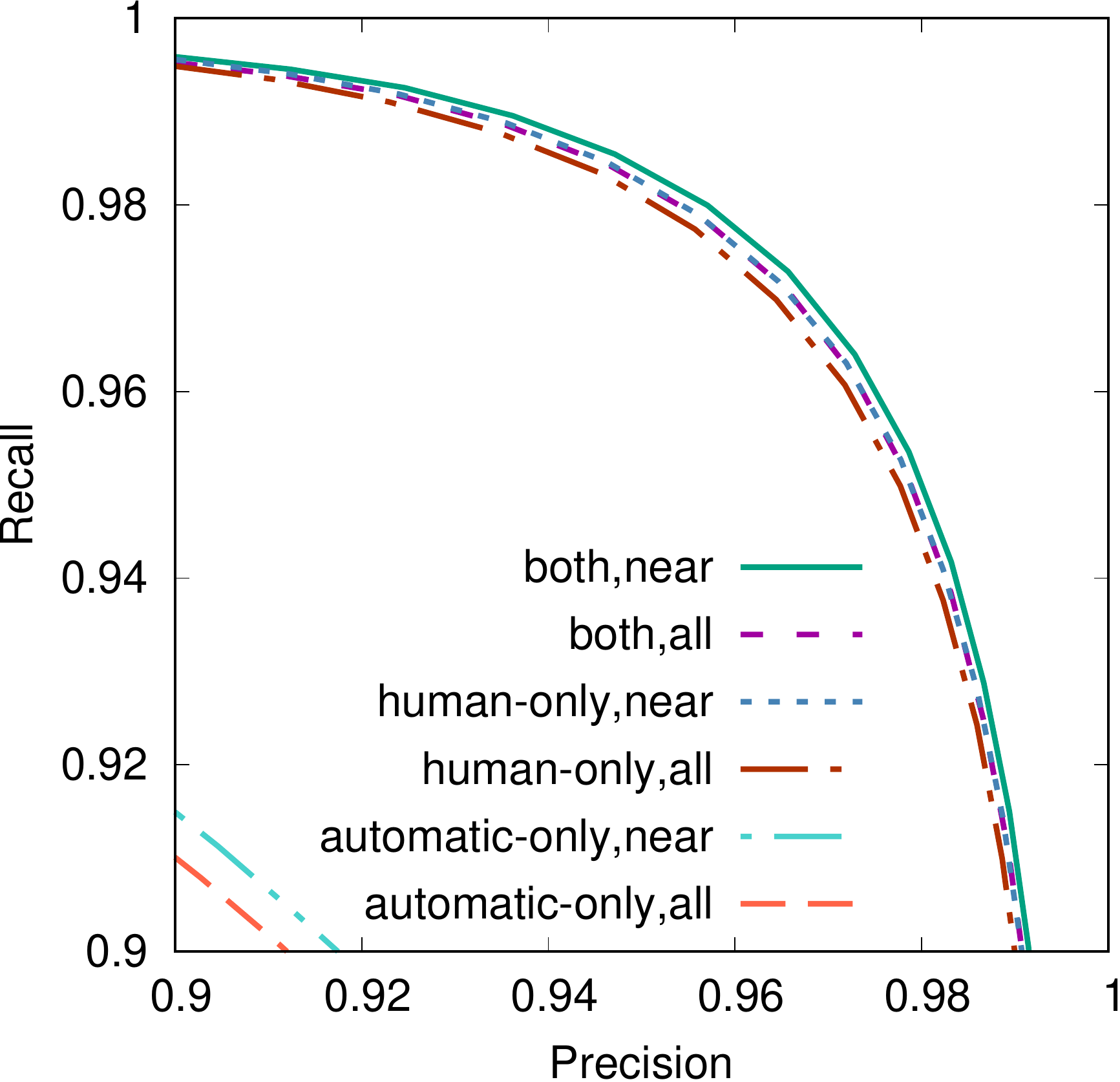}
    }
	\subfloat[\hspace{\subfigLabelPadding}L04-conv-dec]{
    	\includegraphics[width=0.32\linewidth]{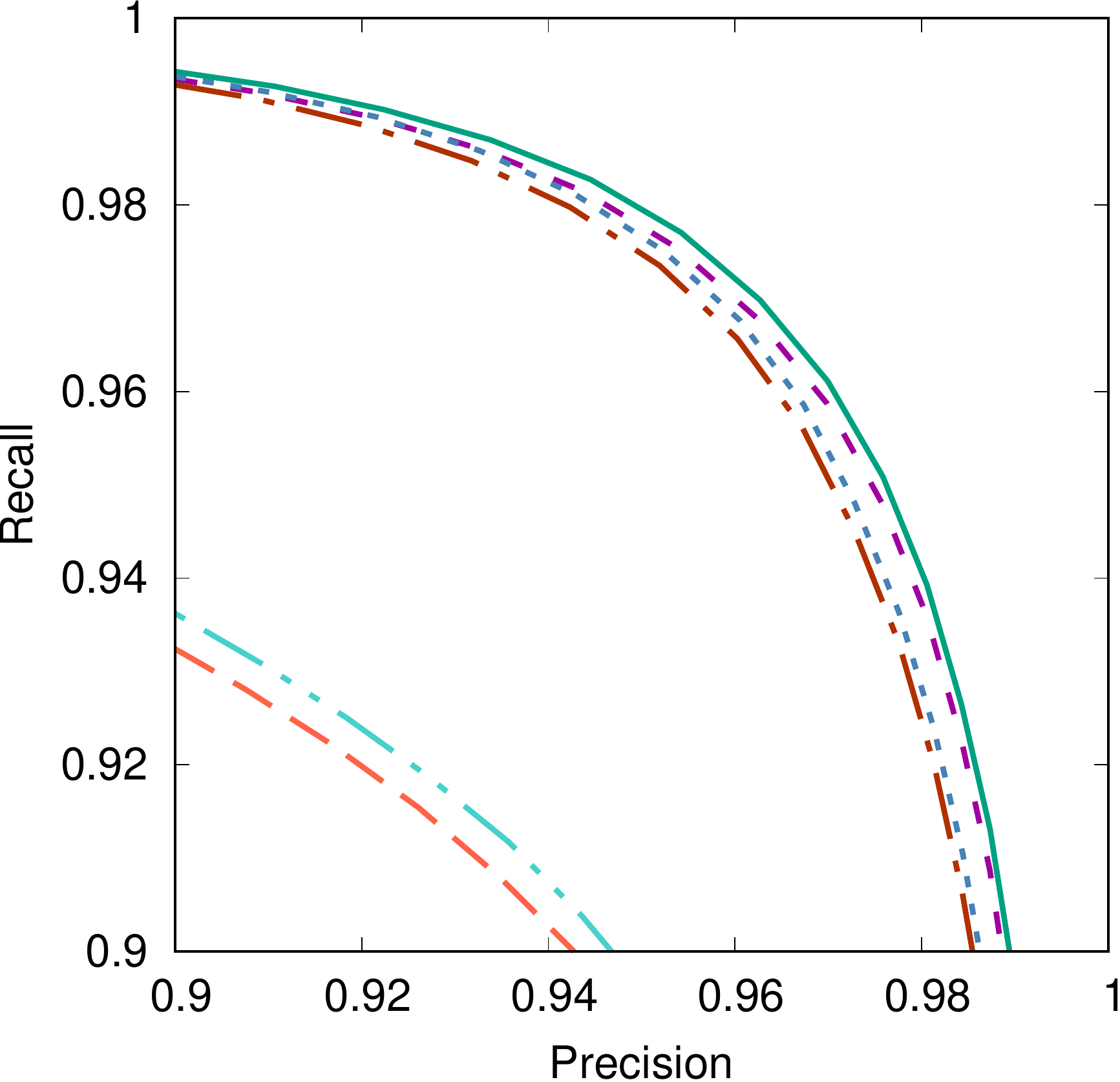}
    }
	\subfloat[\hspace{\subfigLabelPadding}L03+deconv-inc]{
    	\includegraphics[width=0.32\linewidth]{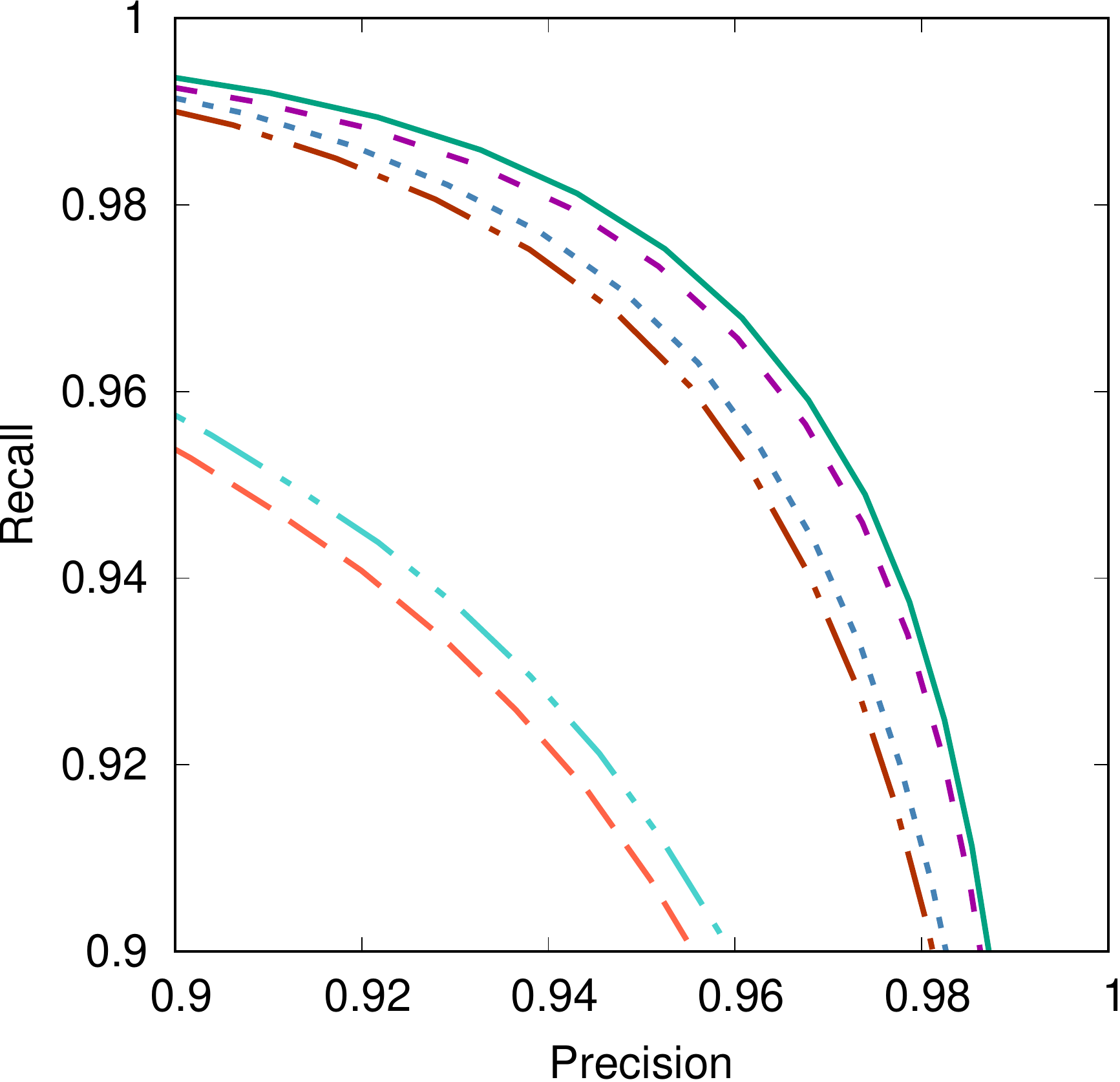}
    }
    \caption{Comparison of the different CNN topologies. The networks were trained using the human-made annotations (label \emph{human-only}), artificial annotations (\emph{automatic-only}) and both datasets for initialization and training (label \emph{both}). \emph{All} the points of input LiDAR frame were processed, or only the points within the $60$m range (label \emph{near}) were taken into the account.}
    \label{fig:pr-different-evals}
    \vspace{-1em}
\end{figure*}

\begin{figure*}[t]
	\centering
	\subfloat{
    	\includegraphics[width=0.49\linewidth]{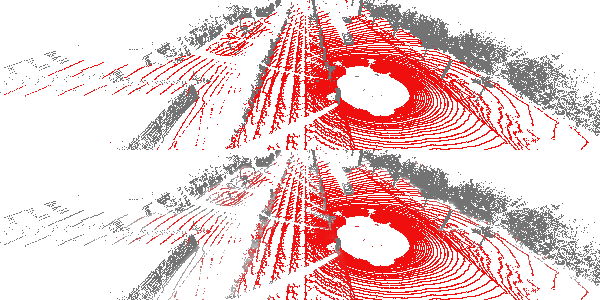}
    }
	\subfloat{
    	\includegraphics[width=0.49\linewidth]{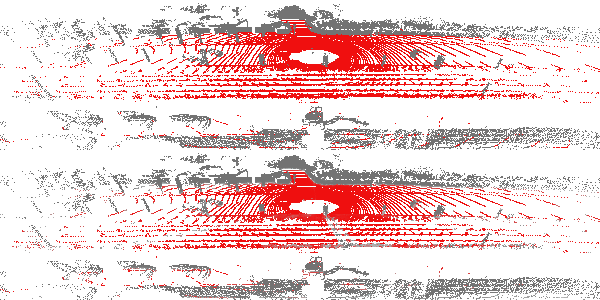}
    }
    \caption{\approve{Ground segmentations (outputs of CNN \emph{L05+deconv}, bottom) for different LiDAR scans compared with human-made annotations (up). The results are near ideal but the differences are still visible under closer inspection.}}
    \label{fig:gt-vs-cnn}
    \vspace{-1em}
\end{figure*}

In \Figure{fig:pr-different-evals}, the precision-recall curves of different network topologies trained and evaluated in different ways are shown. We compared CNNs which were trained either by using the human-made annotations only \approve{(label \emph{human-only})}, or just automatically annotated dataset \approve{(\emph{automatic-only})}, or using both datasets together \approve{(label \emph{both})}. Moreover, we evaluated the accuracy when all points are considered \approve{(label \emph{all})}, or when the maximal range is limited \approve{to $60$m (label \emph{near}) as used also by Zhang \cite{Zhang15}. The examples of CNN outputs can be found in \Figure{fig:gt-vs-cnn}.}

The results depicted in \Figure{fig:pr-different-evals} show that cases where reasoning about the ground was made only within the certain range (label \emph{near}) yield better results. This is expected, since the density of measurements in farther areas is much lower. Also, the CNNs trained with human annotated datasets behave more accurately than CNNs trained on artificial data (evaluation is always made using the human annotations). An interesting fact is that this gap is less significant for networks with smaller architectures (e.g. \cnn{L03} compared to \cnn{L05}). This is probably caused by higher generalization which compromises discriminative power when learned on real annotations.

\Table{tab:time-performance} shows the average processing time of proposed networks using CPU only implementation (Intel i5-6500) and using GPU acceleration (GeForce GTX 770) on a standard desktop computer. These numbers indicate the usability of the networks for certain mobile robot platforms. \cnn{L03+decov-inc} requires low CPU consumption and therefore it is suitable also for small robots with low computational power. On the contrary, the \cnn{L05+deconv} topology would be suitable for platforms where GPU acceleration is available because of the superior accuracy. 

As was said before, the main advantage of our method is superior time performance when compared to the method of Zhang et al.~\cite{Zhang15}. In our experiments, when using the Zhang's MATLAB implementation, the processing time of Velodyne HDL-$64$E LiDAR frame was $145$\,sec and consumed $11$\,GB of memory on average (note: no memory swapping which would compromise the performance happened during the experiments). Also, when we decreased the maximal range (and also the size of the internal $3$D polar grid) to $30$m, the processing time dropped to $75$sec per frame and the memory consumption to approximately one half. However, this is still really far from real-time performance. 

\addtolength{\textheight}{-0.1cm}   

\section{Conclusion}

We presented a real time and robust ground segmentation method of Velodyne LiDAR data, which outperforms the current state-of-the art methods in both the accuracy and speed. 
\approve{Our results show that the sparse LiDAR data can be encoded into the dense $2$D representation and processed by CNN}.
Our method improved the precision of state of the art \cite{Zhang15} (by $0.5\%$) and significantly improved the speed of ground segmentation process from minutes to $140\,\mathrm{ms}$ using CPU and $7\,\mathrm{ms}$ with GPU acceleration.

\approve{We showed that CNN approach is suitable for simpler task of ground segmentation where the results are near ideal. In the follow-up work, we want to explore the potential of this approach in more challenging semantic segmentation or move detection and also in quite different tasks of visual odometry estimation in the LiDAR data or point cloud registration.}

As a secondary outcome of our work, we created the dataset with ground annotated and make it publicly available along with the annotation tool. Such data can be used for designing, training, and evaluation of other ground segmentation approaches.

\bibliographystyle{bib/IEEEtran}
\bibliography{bib/bibliography}

\end{document}